\documentclass[conference,letterpaper]{IEEEtran}

\usepackage[utf8]{inputenc} 
\usepackage[T1]{fontenc}    
\usepackage{hyperref}       
\usepackage{url}            
\usepackage{booktabs}       
\usepackage{amsfonts}       
\usepackage{nicefrac}       
\usepackage{microtype}      
\usepackage{amsmath}
\usepackage{amssymb}
\usepackage{multirow}
\usepackage{array}
\usepackage{graphicx}
\usepackage{siunitx}
\usepackage{footnote}
\usepackage{wrapfig}
\usepackage{flushend}

\newcommand{\argmax}{\operatornamewithlimits{arg\,max}}
\DeclareSIUnit\images{images}
\DeclareSIUnit\fps{fps}

\newcommand*{\affaddr}[1]{#1} 
\newcommand*{\affmark}[1][*]{\textsuperscript{#1}}
\newcommand*{\email}[1]{\texttt{#1}}

\title{\Large\bfseries Ternary Neural Networks for Resource-Efficient AI Applications}

\author{%
Hande~Alemdar\affmark[1], Vincent~Leroy\affmark[1], Adrien~Prost-Boucle\affmark[2], Frédéric~Pétrot\affmark[2]\\
\affaddr{\affmark[1] Univ. Grenoble Alpes, CNRS, Grenoble INP, LIG, F-38000 Grenoble, France}\\
\affaddr{\affmark[2]Univ. Grenoble Alpes, CNRS, Grenoble INP, TIMA, F-38000 Grenoble, France}\\
\email{\{name.surname\}@univ-grenoble-alpes.fr}
}

\date{}

\begin{document}

\maketitle

\begin{abstract}
The computation and storage requirements for Deep Neural Networks (DNNs) are usually high.
This issue limits their deployability on ubiquitous computing devices such as smart phones, wearables and autonomous drones.
In this paper, we propose ternary neural networks (TNNs) in order to make deep learning more resource-efficient.
We train these TNNs using a teacher-student approach based on a novel, layer-wise greedy methodology.
Thanks to our two-stage training procedure, the teacher network is still able to use state-of-the-art methods such as dropout and batch normalization to increase accuracy and reduce training time.
Using only ternary weights and activations, the student ternary network learns to mimic the behavior of its teacher network without using any multiplication.
Unlike its \{-1,1\} binary counterparts, a ternary neural network inherently prunes the smaller weights by setting them to zero during training.
This makes them sparser and thus more energy-efficient.
We design a purpose-built hardware architecture for TNNs and implement it on FPGA and ASIC.
We evaluate TNNs on several benchmark datasets and demonstrate up to 3.1$\times$ better energy efficiency with respect to the state of the art while also improving accuracy. 
\end{abstract}

\section{Introduction}

Deep neural networks (DNNs) have achieved state-of-the-art results on a wide range of AI tasks including computer vision~\cite{Hertel2015}, speech recognition~\cite{Graves2013} and natural language processing~\cite{bahdanau2014neural}.
As DNNs become more complex, their number of layers, number of weights, and computational cost increase.
While DNNs are generally trained on powerful servers with the support of GPUs, they can be used for classification tasks on a variety of hardware.
However, as the networks get bigger, their deployability on autonomous mobile devices such as drones and self-driving cars and mobile phones diminishes due to the extreme hardware resource requirements imposed by high number of synaptic weights and floating point multiplications.
Our goal in this paper is to obtain DNNs that are able to classify at a high throughput on low-power devices without compromising too much accuracy.

In recent years, two main directions of research have been explored to reduce the cost of DNNs classifications.
The first one preserves the floating point precision of DNNs, but drastically increases sparsity and weights sharing for compression~\cite{han2015deep,EIE2016}.
This has the advantage of significantly diminishing memory and power consumption while preserving accuracy.
However, the power savings are limited by the need for floating-point operation.
The second direction reduces the need for floating-point operations using weight discretization~\cite{CourbariauxB16,Dorefa16,LiL16,lin2015neural}, with extreme cases such as binary neural networks completely eliminating the need for multiplications~\cite{Kim2015,esser2015backpropagation,Esser2016}.
The main drawback of these approaches is a significant degradation in the classification accuracy in return for a limited gain in resource efficiency.

This paper introduces ternary neural networks (TNNs) to address these issues and makes the following contributions:
\begin{itemize}
\item We propose a teacher-student approach for obtaining Ternary NNs with weights and activations constrained to $\{-1,0,1\}$.
The teacher network is trained with stochastic firing using back-propagation, and can benefit from all techniques that exist in the literature such as dropout~\cite{srivastava2014dropout}, batch normalization~\cite{ioffe2015batch}, and convolutions,
The student network has the same architecture and, for each neuron, mimics the behavior of the equivalent neuron in the teacher network without using any multiplications,
\item We design a specialized hardware that is able to process TNNs at up to 2.7$\times$ better throughput, 3.1$\times$ better energy efficiency and 635$\times$ better area efficiency than state-of-the-art and with competitive accuracy,
\item We make the training code publicly available~\footnote{https://github.com/slide-lig/tnn-train} and provide a demonstration hardware design for TNNs using FPGA.~\footnote{http://tima.imag.fr/sls/research-projects/tnn-fpga-implementation/}
\end{itemize}

The rest of this paper is organized as follows.
In the following section, we introduce our procedure for training the ternary NNs detailing our use of teacher-student paradigm to eliminate the need for multiplications altogether during test time, while still benefiting all state-of-the-art techniques such as batch normalization and dropout during training.
In Section~\ref{sec:relwork}, we provide a survey of related works that we compare our performance with.
We present our experimental evaluation on ternarization and the classification performance on five different benchmark datasets in Section~\ref{sec:exp}.
In Section~\ref{sec:hw}, we describe our purpose-built hardware that is able to handle both fully connected multi-layer perceptrons (MLPs) and convolutional NNs (CNNs) with a high throughput and a low-energy budget.
Finally, we conclude with a discussion and future studies in Section~\ref{sec:concl}.

\section{Training Ternary Neural Networks}
We use a two-stage teacher-student approach for obtaining TNNs.
First, we train the teacher network with stochastically firing ternary neurons.
Then, we let the student network learn how to imitate the teacher's behavior using a layer-wise greedy algorithm.
Both the teacher and the student networks have the same architecture.
The student network's weights are the ternarized version of the teacher network's weights.
The student network uses a step function with two thresholds as the activation function.
In Table~\ref{tab:definitions}, we provide our notations and descriptions. 
In order to emphasize the difference, we denote the discrete valued parameters and inputs with a bold font.
Real-valued parameters are denoted by normal font. 
We use $[.]$ to denote a matrix or a vector. 
We use subscripts for enumeration purposes and superscripts for differentiation.
$\mathbf{n_i^t}$  is defined as the output of neuron $i$ in teacher network and $\mathbf{n_i^s}$ is the output of neuron $i$ in the student network.
We detail the two stages in the following subsections.

\begin{table}[t!]
\centering
\caption{Ternary Neural Network Definitions for a Single Neuron $i$}\label{tab:definitions}
\scriptsize
\begin{tabular}{@{}l@{}l@{}l@{}}
\toprule
 & \textbf{Teacher network} & \textbf{Student Network} \\
\midrule
Weights & $W_i = [ w_j ] ,  w_j  \in \mathbb{R} $ & $\mathbf{W_i} = [ \mathbf{w_j} ] , \mathbf{w_j} \in \{-1,0,1\} $ \\
\hline
Bias & $b_i \in \mathbb{R} $ &  $ \mathbf{b_i}^{\mathit{lo}} \in \mathbb{Z} $  \\
	& 						&  $ \mathbf{b_i}^{\mathit{hi}} \in \mathbb{Z}$ \\
\hline
Transfer &  $ y_i = W_i^\intercal\mathbf{x} + b_i  $ & $ \mathbf{y_i} = \mathbf{W_i}^\intercal\mathbf{x}$ \\
Function &  &  \\
\hline
Act. Fun  $\quad$ &
$ \mathbf{n_i^t} =
	\begin{cases}
		-1  \quad  \text{with prob.  $-\rho$ if $\rho < 0$}  \\
		1 	\qquad  \text{with prob.  $\rho$ if $\rho > 0$}  \\
		0    \qquad  \text{otherwise} \end{cases}$
 & $\mathbf{n_i^s} = \begin{cases}
		-1  	 \quad \text{if $\mathbf{y_i} < \mathbf{b_i}^{\mathit{lo}} $}\\
		1 	 \qquad \text{if $\mathbf{y_i}> \mathbf{b_i}^{\mathit{hi}}$} \\
		0    \qquad \text{otherwise}	\end{cases}$ \\
 & where  $\rho = tanh(y_i),  \rho\in(-1,1) $  &  \\
\bottomrule
\end{tabular}
\end{table}

\subsection{The Teacher Network}
The teacher network can have any architecture with any number of neurons, and can be trained using any of the standard training algorithms.
We train the teacher network with a single constraint only:
it has stochastically firing ternary neurons with output values of $-1$, $0$, or $1$.
The benefit of this approach is that we can use any technique that already exists for efficient NN training, such as batch normalization~\cite{ioffe2015batch}, dropout~\cite{srivastava2014dropout}, etc.
In order to have a ternary output for teacher neuron $i$ denoted as $\mathbf{n_i^t}$, we add a stochastic firing step after the activation step.
For achieving this stochastically, we use \emph{tanh} (hyperbolic tangent), \emph{hard tanh}, or \emph{soft-sign} as the activation function of the teacher network so that the neuron output has $(-1,1)$ range before ternarization.
We use this range to determine the ternary output of the neuron as described in Table~\ref{tab:definitions}.
Although we do not require any restrictions for the weights of the teacher network, several studies showed that it has a regularization effect and reduces over-fitting~ \cite{LiL16,lin2015neural}.
Our approach is compatible with such a regularization technique as well.

\subsection{The Student Network}
After the teacher network is trained, we begin the training of the student network.
The goal of the student network is to predict the output of the teacher real-valued network.
Since we use the same architecture for both networks, there is a one-to-one correspondence between the neurons of both.
Each student neuron denoted as $\mathbf{n_i^s}$ learns to mimic the behavior of the corresponding teacher neuron $\mathbf{n_i^t}$ individually and independently from the other neurons.
In order to achieve this, a student neuron uses the corresponding teacher neuron's weights as a guide to determine its own ternary weights using two thresholds $t_i^\mathit{lo}$ (for the lower threshold) and $t_i^{\mathit{hi}}$ (for the higher one) on the teacher neuron's weights.
This step is called the \textit{weight ternarization}.
In order to have a ternary neuron output, we have a step activation function of two thresholds $\mathbf{b_i}^{\mathit{lo}}$ and $\mathbf{b_i}^{\mathit{hi}}$.
The \textit{output ternarization} step determines these.

\begin{figure}[t!]
\centering
\includegraphics[width=\linewidth]{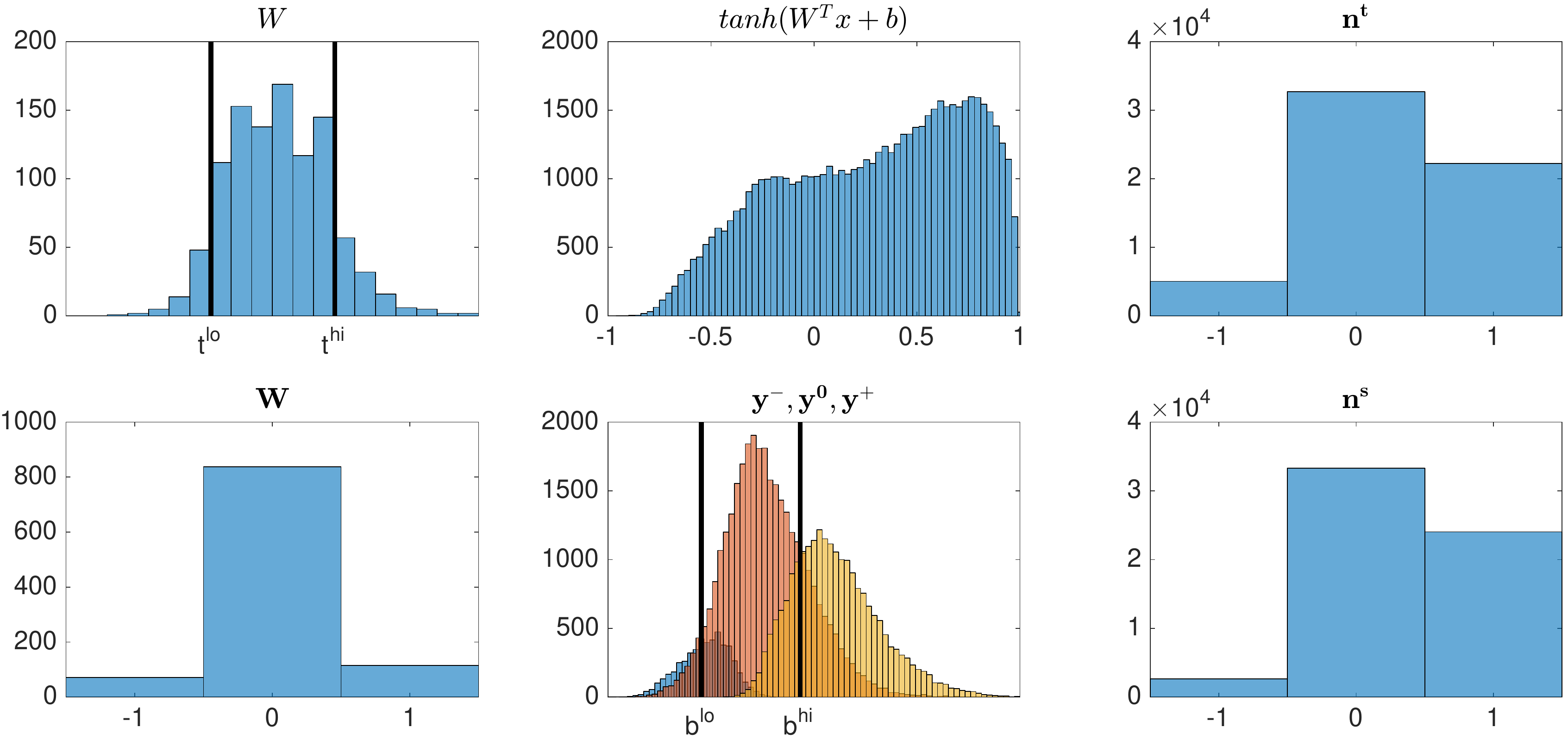}
\caption{Example ternarization for a single neuron}
\label{fig:neruon}
\end{figure}

Figure~\ref{fig:neruon} depicts the ternarization procedure for a sample neuron.
In the top row, we plot the distributions of the weights, activations and ternary output of a sample neuron in the teacher network respectively.
The student neuron's weight distribution that is determined by $t_i^\mathit{lo}$ and $t_i^{\mathit{hi}}$ is plotted below the teacher's weight distribution.
We use the transfer function output of the student neuron, grouped according to the teacher neuron's output on the same input, to determine the thresholds for the step activation function.
In this way, the resulting output distribution for both the teacher and the student neurons are similar.
In the following subsections we detail each step.

\subsubsection{Output Ternarization}
The student network uses a two-thresholded step activation function to have ternary output as described in Table~\ref{tab:definitions}.
Output ternarization finds the step activation function's thresholds $\mathbf{b_i}^{\mathit{lo}}$ and $\mathbf{b_i}^{\mathit{hi}}$, for a ternary neuron $i$, for a given set of ternary weights $\mathbf{W}$.
In order to achieve this, we compute three different transfer function output distributions for the student neuron, using the teacher neuron's ternary output value on the same input.
We use $\mathbf{y^-}$ to denote the set of transfer function outputs of the student neuron for which the teacher neuron's output value is $-1$.
$\mathbf{y^0}$ and $\mathbf{y^+}$ are defined in the same way for teacher neuron output values $0$ and $1$, respectively.

We use a simple classifier to find the boundaries between these three clusters of student neuron transfer function outputs, and use the boundaries as the two thresholds $\mathbf{b_i}^{\mathit{lo}}$ and $\mathbf{b_i}^{\mathit{hi}}$ of the step activation function.
The classification is done using a linear discriminant on the kernel density estimates of the three distributions.
The discriminant between $\mathbf{y^+}$ and $\mathbf{y^0}$ is selected as the $\mathbf{b_i}^{\mathit{hi}}$, and the discriminant between $\mathbf{y^-}$ and $\mathbf{y^0}$ gives the $\mathbf{b_i}^{\mathit{lo}}$.

\subsubsection{Weight Ternarization} 
During weight ternarization, the order and the sign of the teacher network's weights are preserved. 
We ternarize the weights of the $i^{th}$ neuron of the teacher network using two thresholds $ t_i^\mathit{lo}$ and $t_i^{\mathit{hi}}$ such that $min( W_i ) \leqslant t_i^\mathit{lo} \leqslant 0$ and $0  \leqslant t_i^\mathit{hi} \leqslant  max( W_i )$. The weights for the $i^{th}$ student neuron are obtained by weight ternarization as follows 
\begin{equation}
  ternarize( W_i | t_i^\mathit{lo}, t_i^\mathit{hi} ) = \mathbf{W_i} = [  \mathbf{w_j} ]
\end{equation}
where 
\begin{equation}
\mathbf{w_j} = \begin{cases}
	 \begin{split}
		-1  	& \quad \text{if $w_j < t_i^\mathit{lo} $}\\
		0   & \quad \text{if $t_i^\mathit{lo} \geqslant  w_j \geqslant  t_i^{\mathit{hi}} $} \\
		1 	& \quad \text{if $w_j > t_i^{\mathit{hi}} $} \\
	 \end{split}
	\end{cases}
\end{equation}
We find the optimal threshold values for the weights by evaluating the ternarization quality with a score function.
For a given neuron with $p$ positive weights and $n$ negative weights, the total number of possible ternarization schemes is $np$ since we respect the original sign and order of weights. 
For a given configuration of the positive and negative threshold values $t^\mathit{hi}$ and $t^\mathit{lo}$ for a given neuron, we calculate the following score for assessing the performance of the ternary network, mimicking the original network.
\begin{equation}
S_{t^\mathit{lo},t^\mathit{hi}} = \sum_d p(\mathbf{n^t}=\pm1 | \mathbf{x^t_d} )^{\mathbb{I}(\mathbf{n^s}=\pm1 | \mathbf{x^s_d})} p(\mathbf{n^t}=0 | \mathbf{x^t_d} )^{\mathbb{I}(\mathbf{n^s}=0 | \mathbf{x^s_d})}
\end{equation}
where $\mathbf{n^t}$ and $\mathbf{n^s}$ denote the output of the teacher neuron and student neuron, respectively.

$\mathbf{x^t_d}$ is the $d^{th}$ input sample for the teacher neuron, and $\mathbf{x^s_d}$ is the input  $d^{th}$ input sample for the student neuron.
Note that $\mathbf{x^t_d} \neq \mathbf{x^s_d}$ after the first layer.
Since we ternarize the network in a feed-forward manner, in order to prevent ternarization errors from propagating to upper layers, we always use the teacher's original input to determine its output probability distribution.
The output probability distribution for the teacher neuron for input $d$, $p(\mathbf{n^t} | \mathbf{x^t_d} )$, is calculated using stochastic firing as described in Table~\ref{tab:definitions}.
The output probability distribution for the student neuron for input $d$, $p(\mathbf{n^s} | \mathbf{x^s_d} )$ is calculated using the ternary weights $\mathbf{W}$ with the current configuration of $t^\mathit{lo}$, $t^\mathit{hi}$, and the step activation function thresholds. 
These thresholds, $\mathbf{b}^{\mathit{hi}}$ and $\mathbf{b}^{\mathit{lo}}$ are selected according to the current ternary weight configuration $\mathbf{W}$.

The output probability values are accumulated as scores over all input samples only when the output of the student neuron matches the output of the teacher neuron.
The optimal ternarization of weights is determined by selecting the configuration with the maximum score.
\begin{equation}
\mathbf{W}^*= \argmax_{ t^\mathit{lo},t^\mathit{hi} } S_{ t^\mathit{lo},t^\mathit{hi} }
\end{equation}

The worst-case time complexity of the algorithm is $O(\Vert W \Vert^2)$.
We propose using a greedy dichotomic search strategy instead of a fully exhaustive one.
We make a search grid over $n$ candidate values for $t^\mathit{lo}$ by $p$ values for $t^\mathit{hi}$.
We select two equally-spaced pivot points along one of the dimensions, $n$ or $p$.
Using these pivot points, we calculate the maximum score along the other axis.
We reduce the search space by selecting the region in which the maximum point lies.
Since we have two points, we reduce the search space to two-thirds at each step.
Then, we repeat the search procedure in the reduced search space.
This faster strategy runs in $O(log^2 \Vert W \Vert )$, and when there are no local maxima it is guaranteed to find the optimal solution.
When there are multiple local extremum, it may get stuck.
Fortunately, we can detect the possible sub-optimal solutions, using the score values obtained for the student neuron.
By using a threshold on the output score for a student neuron, we can selectively use exhaustive search on a subset of neurons.
Empirically, we find these cases to be rare.
We provide a detailed analysis in Section~\ref{sec:tern_perf}.

The ternarization of the output layer is slightly different since it is a soft-max classifier.
In the ternarization process, instead of using the teacher network's output, we use the actual labels in the training set.
Again, we treat neurons independently but we make several iterations over each output neuron in a round-robin fashion.
After each iteration we check against convergence.
In our experiments, we observed that the method converges after a few passes over all neurons.

Our layer-wise approach allows us to update the weights of the teacher network before ternarization of any layer.
For this optional weight update, we use a staggered retraining approach in which only the non-ternarized layers are modified.
After the teacher network's weights are updated, input to a layer for both teacher and student networks become equal, $\mathbf{x^t_d}=\mathbf{x^s_d}$.
We use early stopping during this optional retraining and we find that a few dozen of iterations suffice.

\section{Related Work}
\label{sec:relwork}

In this section, we give a brief survey on several related works in energy-efficient NNs.
In Table~\ref{tab:comparison}, we provide a comparison between our approach and the related works that use binary or ternary weights in the deployment phase by summarizing the constraints put on inputs, weights and activations during training and testing.

\begin{table*}[t!]
\label{tab:comparison}
\centering
\caption{Comparison of several approaches for resource-efficient neural networks}
\scriptsize
\begin{tabular}{@{}l lll c lll@{}}
\toprule
& \multicolumn{3}{c}{\textbf{Training}}&& \multicolumn{3}{c}{\textbf{Deployment}} \\
\cmidrule{2-4} \cmidrule{6-8}
\textbf{Method} & \textbf{Inputs} & \textbf{Weights} & \textbf{Activations} && \textbf{Inputs} & \textbf{Weights} & \textbf{Activations} \\
\midrule
BC \cite{courbariaux2015binaryconnect}, TC \cite{lin2015neural}, TWN  \cite{LiL16}  & $\mathbb{R}$ & $\{-1,0,1\}$ &$\mathbb{R}$  &&$\mathbb{R}$& $\{-1,0,1\}$ &$\mathbb{R}$  \\
\hline
Binarized NN \cite{CourbariauxB16} &$\mathbb{R}$& $\{-1,1\}$  & $\{-1,1\}$  && $\mathbb{R}$ & $\{-1,1\}$ & $\{-1,1\}$ \\
\hline
XNOR-Net\cite{rastegariECCV16} &$\mathbb{R}$& $\{-1,1\}$  & $\{-1,1\}$ with $K,\alpha \in \mathbb{R} $   && $\mathbb{R}$ & $\{-1,1\}$ & $\{-1,1\}$ with $K,\alpha \in \mathbb{R} $ \\
\hline
EBP\cite{soudry2014expectation} & $\mathbb{R}$ & $\mathbb{R}$ & $\mathbb{R}$  && $\mathbb{R}$ & $\{-1,0,1\}$ & $\{-1,1\}$ \\
\hline
Bitwise NN \cite{Kim2015} & $(-1,1)$ & $(-1,1)$ &  $(-1,1)$   && $\{-1,1\}$ & $\{-1,0,1\}$  & $\{-1,1\}$  \\
	    				     & $[0,1]$  & $(-1,1)$ &  $(-1,1)$   && $\{0,1\}$  & $\{-1,0,1\}$  &  $\{-1,1\}$ \\
\hline
TrueNorth \cite{esser2015backpropagation} & $[0,1]$ & $[-1,1]$ & $[0,1]$  && $\{0,1\}$ & $\{-1,0,1\}$ & $\{0,1\}$ \\
\hline
\textbf{TNN (This Work)} & $\{-1,0,1\}$& $\mathbb{R}$ &  $\{-1,0,1\}$ && $\{-1,0,1\}$& $\{-1,0,1\}$ & $\{-1,0,1\}$ \\
\bottomrule
\end{tabular}
\end{table*}

Courbariaux et al. \cite{courbariaux2015binaryconnect} propose the BinaryConnect (BC) method for binarizing only the weights, leaving the inputs and the activations as real-values.
The same idea is also used as TernaryConnect (TC) in \cite{lin2015neural} and Ternary Weight Networks (TWN) in \cite{LiL16} with ternary $\{-1,0,1\}$ weights instead of binary $\{-1,1\}$.
They use the back-propagation algorithm with an additional weight binarization step.
In the forward pass, weights are binarized either deterministically using their sign, or stochastically.
Stochastic binarization converts the real-valued weights to probabilities with the use of the hard-sigmoid function, and then decides the final value of the weights with this.
In the back-propagation phase, a quantization mechanism is used so that the multiplication operations are converted to bit-shift operations.
While this binarization scheme helps reducing the number of multiplications during training and testing, it is not fully hardware-friendly since it only reduces the number of floating point multiplication operations.
Recently, the same idea is extended to the activations of the neurons also \cite{CourbariauxB16}.
In Binarized NN, the sign activation function is used for obtaining binary neuron activations.
Also, shift-based operations are used during both training and test time in order to gain energy-efficiency.
Although this method helps improving the efficiency in multiplications it does not eliminate them completely.

XNOR-Nets \cite{rastegariECCV16} provide a solution to convert convolution operations in CNNs to bitwise operations.
The method first learns a discrete convolution together with a set of real-valued scaling factors ($K,\alpha \in \mathbb{R} $).
After the convolution calculations are handled using bit-wise operations, the scaling factors are applied to obtain actual result.
This approach is very similar to Binarized NN and helps reducing the number of floating point operations to some extent.

Following the same goal, DoReFa-Net \cite{Dorefa16} and Quantized Neural Networks (QNN) \cite{QNN2016} propose using $n$-bit quantization for weights, activations as well as gradients.
In this way, it is possible to gain speed and energy efficiency to some extent not only during training but also during inference time.
Han et al. \cite{han2015deep} combine several techniques to achieve both quantization and compression of the weights by setting the relatively unimportant ones to 0.
They also develop a dedicated hardware called Efficient Inference Engine (EIE) that exploits the quantization and sparsity to gain large speed-ups and energy savings, only on fully connected layers currently \cite{EIE2016}.

Soudry et al. \cite{soudry2014expectation} propose Expectation Backpropagation (EBP), an algorithm for learning the weights of a binary network using a variational Bayes technique.
The algorithm can be used to train the network such that, each weight can be restricted to be binary or ternary values.
The strength of this approach is that the training algorithm does not require any tuning of hyper-parameters, such as learning rate as in the standard back-propagation algorithm.
Also, the neurons in the middle layers are binary, making it hardware-friendly.
However, this approach assumes the bias is real and it is not currently applicable to CNNs.

All of the methods described above are only partially discretized, leading only to a reduction in the number of floating point multiplication operations.
In order to completely eliminate the need for multiplications which will result in maximum resource efficiency, one has to discretize the network completely rather than partially.
Under these extreme limitations, only a few studies exist in the literature.

Kim and Smaragdis propose Bitwise NN \cite{Kim2015} which is a completely binary approach, where all the inputs, weights, and the outputs are binary.
They use a straightforward extension of back-propagation to learn bitwise network's weights.
First, a real-valued network is trained by constraining the weights of the network using $tanh$.
Also $tanh$ non-linearity is used for the activations to constrain the neuron output to $(-1,1)$.
Then, in a second training step, the binary network is trained using the real-valued network together with a global sparsity parameter.
In each epoch during forward propagation, the weights and the activations are binarized using the sign function on the original constrained real-valued parameters and activations.
Currently, CNNs are not supported in Bitwise-NNs.

IBM announced an energy efficient TrueNorth chip, designed for spiking neural network architectures \cite{merolla2014}.
Esser et al. \cite{esser2015backpropagation} propose an algorithm for training networks that are compatible with IBM TrueNorth chip.
The algorithm is based on backpropagation with two modifications.
First, Gaussian approximation is used for the summation of several Bernoulli neurons, and second, values are clipped to satisfy the boundary requirements of TrueNorth chip.
The ternary weights are obtained by introducing a synaptic connection parameter that determines whether a connection exits.
If the connection exists, the sign of the weight is used.
Recently, the work has been extended to CNN architectures as well~\cite{Esser2016}.

\section{Experimental Assessment of Ternarization and Classification}\label{sec:exp}
The main goals of our experiments are to demonstrate, (i) the performance of the ternarization procedure with respect to the real-valued teacher network, and (ii) the classification performance of fully discretized ternary networks.

We perform our experiments on several benchmarking datasets using both multi-layer perceptrons (MLP) in a permutation-invariant manner and convolutional neural networks (CNN) with varying sizes.
For the MLPs, we experiment with different architectures in terms of depth and neuron count.
We use 250, 500, 750, and 1000 neurons per layer for 2, 3, and 4 layer networks.
For the CNNs, we use the following VGG-like architecture proposed by \cite{courbariaux2015binaryconnect}:

$(2 \times \mathit{nC3}) - MP2 - (2 \times \mathit{2nC3}) - \mathit{MP2} - (2 \times \mathit{4nC3}) - MP2 - (2 \times 8\mathit{nFC}) - \mathit{L2SVM}$

\noindent where $C3$ is a $3 \times 3$ convolutional layer, $\mathit{MP2}$ is a $2 \times 2$ max-pooling layer, $\mathit{FC}$ is a fully connected layer. We use $\mathit{L2SVM}$ as our output layer.
We use two different network sizes with this architecture with $n=64$ and $n=128$.
We call these networks CNN-Small and CNN-Big, respectively.

We perform our experiments on the following datasets:

\textbf{MNIST} database of handwritten digits \cite{lecun:1998} is a well-studied database for benchmarking methods on real-world data.
MNIST has a training set of 60K examples, and a test set of 10K examples of $28 \times 28$ gray-scale images.
We use the last 10K samples of the training set as a validation set for early stopping and model selection.

\textbf{CIFAR-10} and  \textbf{CIFAR-100} \cite{CIFAR} are two color-image classification datasets that contain $32 \times 32$ RGB images.
Each dataset consists of 50K images in training and 10K images in test sets.
In CIFAR-10, the images come from a set of 10 classes that contain airplanes, automobiles, birds, cats, deer, dogs, frogs, horses, ships and trucks.
In CIFAR-100, the number of image classes is 100.

\textbf{SVHN (Street View House Numbers)} \cite{SVHN} consists of $32 \times 32$ RGB color images of digits cropped from Street View images.
The total training set size is 604K examples (with 531K less difficult samples to be used as extra) and the test set contains 26K images.

\textbf{GTSRB (German Traffic Sign Recognition Benchmark Dataset)} \cite{GTSRB} is composed of 51839 images of German road signs in 43 classes.
The images have great variability in terms of size and illumination conditions.
Also, the dataset has unbalanced class frequencies.
The images in the dataset are extracted from 1-second video sequences recorded at 30 fps.
In order to have a representative validation set, we extract 1 track at random per traffic sign for validation.
The number of images in train, validation and test set are 37919, 1290 and 12630 respectively.

In order to allow a fair comparison against related works, we perform our experiments in similar configurations.
On MNIST, we only use MLPs.
We minimize cross entropy loss using stochastic gradient descent with a mini-batch size of 100.
During training we use random rotations up to $\pm 10$ degrees.
We report the test error rate associated with the best validation error rate after 1000 epochs.
We do not perform any preprocessing on MNIST and we use a threshold-based binarization on the input.

For other datasets, we use two CNN architectures: CNN-Small and CNN-Big.
As before, we train a teacher network before obtaining the student TNN.
For the teacher network, we use a modified version of Binarized NN's algorithm \cite{CourbariauxB16} and ternarize the weights during training.
In this way, we obtain a better accuracy on the teacher network and we gain considerable speed-up while obtaining the student network.
Since we already have the discretized weights during the teacher network training, we only mimic the output of the neurons using the step activation function with two thresholds for the student network.
During teacher network training, we minimize squared Hinge loss with \textit{adam} with mini-batches of size 200.
We train at most 1000 epochs and report the test error rate associated with the best validation epoch.
For input binarization, we use the approach described in \cite{Esser2016} with either 12 or 24 (on CIFAR100) transduction filters.
We do not use any augmentation on the datasets.

\subsection{Ternarization Performance}
\label{sec:tern_perf}
\begin{figure}[t!]
\centering
\includegraphics[width=\linewidth]{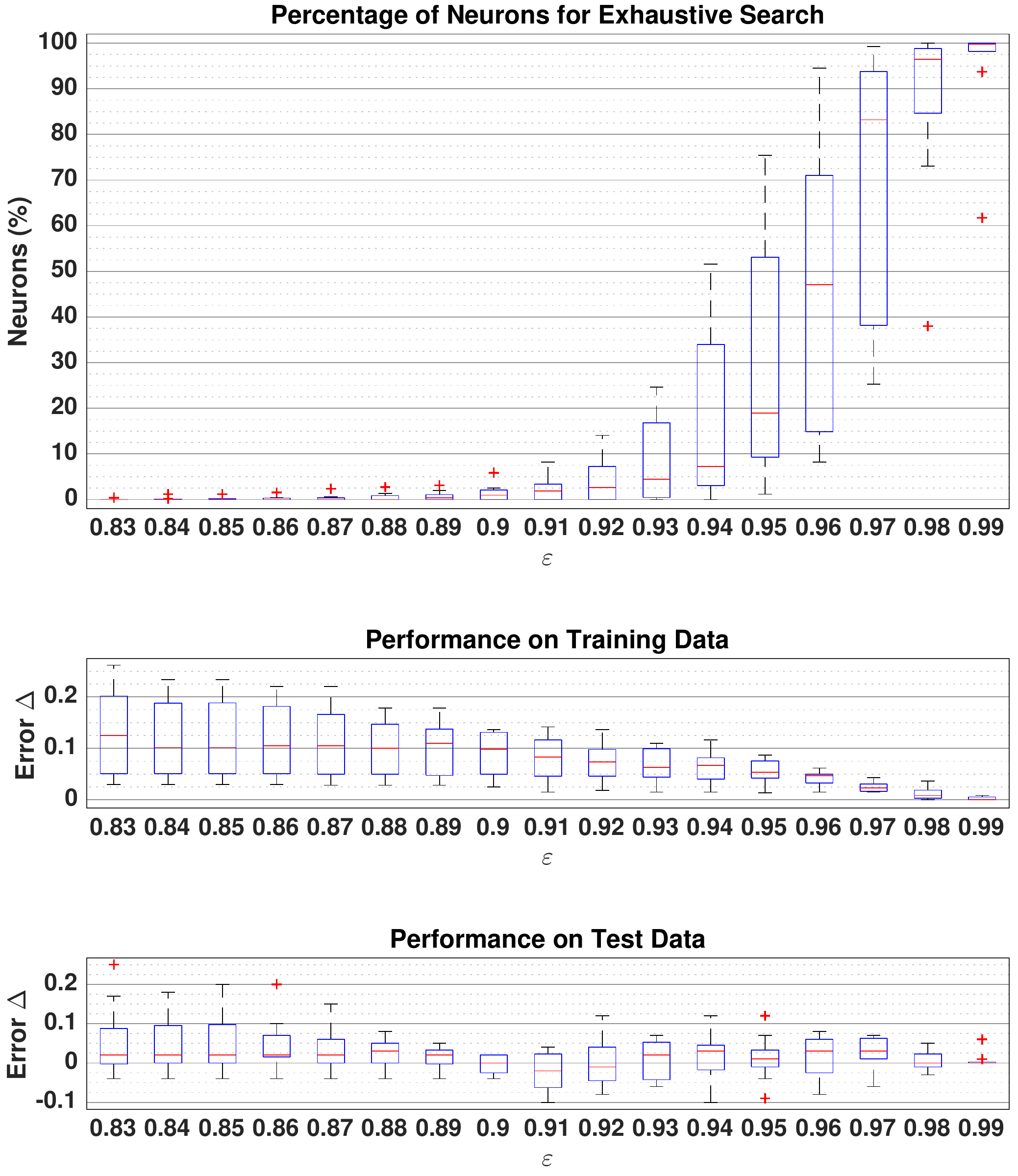}
\caption{The effect of threshold values on run-time and classification performance }\label{fig:sens_analysis}
\end{figure}
The ternarization performance is the ability of the student network to imitate the behavior of its teacher.
We measure this by using the accuracy difference between the teacher network and the student network.
Table~\ref{tab:tern_perf} shows this difference between the teacher and student networks on training and test sets for three different exhaustive search threshold values. $\varepsilon=1$ corresponds to the fully exhaustive search case whereas $\varepsilon=0$ represents fully dichotomic search.
\begin{table}[!b]
\centering
\caption{Accuracy gap due to ternarization for different $\varepsilon$ }\label{tab:tern_perf}
\scriptsize
\begin{tabular}{@{}cc c ll c ll c ll@{}}
\toprule
& & \multicolumn{2}{c}{\boldmath$\varepsilon=1$} && \multicolumn{2}{c}{\boldmath$\varepsilon=0.95$} && \multicolumn{2}{c}{\boldmath$\varepsilon=0$} \\
\cmidrule{3-4}\cmidrule{6-7}\cmidrule{9-10}
\textbf{Neurons} & \textbf{Layers} & \textbf{Train} & \textbf{Test}  && \textbf{Train} & \textbf{Test} && \textbf{Train} & \textbf{Test} \\
\midrule
\multirow{3}{*}{250}&1 & 0.63&0.75&&0.69&0.76&&0.72&0.98\\
&2 & 0.30&0.44&&0.25&0.56&&0.15&0.54\\
&3 & 0.08&0.47&&0.10&0.62&&0.03&0.52\\
\midrule
\multirow{3}{*}{500}&1 & 0.50&0.94&&0.49&0.83&&0.48&0.70\\
&2 & 0.23&0.36&&0.29&0.39&&0.26&0.49\\
&3 & 0.12&0.25&&0.07&0.34&&0.10&0.20\\
\midrule
\multirow{3}{*}{750}&1 & 0.27&0.90&&0.27&0.90&&0.29&1.05\\
&2 & 0.00&0.56&&-0.01&0.65&&0.00&0.67\\
&3 & -0.02&0.43&&-0.03&0.44&&-0.03&0.26\\
\midrule
\multirow{3}{*}{1000}&1 & 0.44&0.66&&0.60&0.87&&0.79&0.95\\
&2 & -0.02&0.59&&-0.07&0.33&&-0.05&0.37\\
&3 & -0.16&0.29&&-0.14&0.34&&-0.14&0.34\\
\bottomrule
\end{tabular}
\end{table}
The results show that the ternarization performance is better for deeper networks.
Since we always use the teacher network's original output as a reference, errors are not amplified in the network.
On the contrary, deeper networks allow the student network to correct some of the mistakes in the upper layers, dampening the errors.
Also, we perform a retraining step with early stopping before ternarizing a layer, since it slightly improves the performance.
The ternarization performance generally decreases with lower $\varepsilon$ threshold values, but the decrease is marginal.
On occasion, performance even increases.
This is due to teacher network's weight update, that allows the network to escape from local minima.
\begin{table}[!b]
\centering
\caption{Classification Performance -  Error Rates (\%) }\label{tab:clas_perf}
\scriptsize
\begin{tabular}{@{}llccccc@{} }
\toprule
&&MNIST&CIFAR10&SVHN&GTRSB&CIFAR100\\
\midrule
\multicolumn{2}{l}{Fully Discretized} & & & & & \\
&\textbf{TNN} (This Work)&1.67&12.11&2.73&0.98&48.40\\
&TrueNorth \cite{esser2015backpropagation,Esser2016}&7.30&16.59&3.34&3.50&44.36\\
&Bitwise NN \cite{Kim2015}&1.33&&&&\\
\midrule
\midrule
\multicolumn{2}{l}{Partially Discretized} & & & & & \\
&Binarized NN \cite{CourbariauxB16}&0.96&10.15&2.53&&\\
&BC \cite{courbariaux2015binaryconnect}&1.29&9.90&2.30&&\\
&TC \cite{lin2015neural}&1.15&12.01&2.42&&\\
&TWN  \cite{LiL16} &0.65&7.44&&&\\
&EBP \cite{Cheng2015}&2.20&&&&\\
&XNOR-Net \cite{rastegariECCV16}&&9.88&&&\\
&DoReFa-Net \cite{Dorefa16}&&&2.40&&\\
\bottomrule
\end{tabular}
\end{table}
In order to demonstrate the effect of $\varepsilon$ in terms of run-time and classification performance, we conduct a detailed analysis without the optional staggered retraining.
Figure~\ref{fig:sens_analysis} shows the distribution of the ratio of neurons that are ternarized exhaustively with different $\varepsilon$, together with the performance gaps on training and test datasets.
The optimal trade-off is achieved with $\varepsilon=0.95$.
Exhaustive search is used for only 20\% of the neurons, and the expected value of accuracy gaps is practically 0.
For the largest layer with 1000 neurons, the ternarization operations take \SI{2}{\minute} and \SI{63}{\minute} for dichotomic and exhaustive search, respectively, on a 40-core Intel(R) Xeon(R) CPU E5-2650 v3 @ 2.30GHz server with 128 GB RAM. For the output layer, the ternarization time is reduced to \SI{21}{\minute} with exhaustive search.

\subsection{Classification Performance}
The classification performance in terms of error rate (\%) on several benchmark datasets is provided in Table~\ref{tab:clas_perf}.
We compare our results to several related methods that we described in the previous section.
We make a distinction between the fully discretized methods and the partially discretized ones because only in the latter, the resulting network is completely discrete and requires no floating points and no multiplications, providing maximum energy efficiency.

Since the benchmark datasets we use are the most studied ones, there are several known techniques and tricks that give performance boosts.
In order to eliminate unfair comparison among methods, we follow the majority's lead and we do not use any data augmentation in our experiments.
Moreover, using an ensemble of classifiers is a common technique for performance boosting in almost all classifiers and is not unique to neural networks \cite{Ciresan:2011b}.
For that reason, we do not use an ensemble of networks and we cite the compatible results in other related works.

MNIST is by far the most studied dataset in deep learning literature.
The state-of-the-art is already down to 21 erroneous classifications ($0.21\%$) which is extremely difficult to obtain without extensive data augmentation.
TNN's error rate on MNIST is $1.67\%$ with a single 3-layer MLP with 750 neurons in each layer.
Bitwise NNs~\cite{Kim2015} with 1024 neurons in 3 layers achieves a slightly better performance.
TNN with an architecture that has similar size to Bitwise NN is worse due to over-fitting.
Since TNN selects a different sparsity level for each neuron, it can perform better on smaller networks, and larger networks cause over-fitting on MNIST.
Bitwise NN's global sparsity parameter has a better regularization effect on MNIST for relatively bigger networks.
Its performance with smaller networks or on other datasets is unknown.
TrueNorth~\cite{esser2015backpropagation} with a single network achieves only $7.30\%$ error rate.
To alleviate the limitations of single network performance, a committee of networks can be used, reducing the error rate to $0.58\%$ with 64 networks.

The error rate of TNN on CIFAR10 is $12.11\%$.
When compared to partially discretized alternatives, a fully discretized TNN is obtained at the cost a few points in the accuracy and exceeds the performance of TrueNorth by more than $4\%$.
On SVHN, it has a similar achievement with lower margins.
For CIFAR100, on the other hand, it does not perform better than TrueNorth.
Given the relatively lower number of related works that report results on CIFAR100 as opposed to CIFAR10, we can conclude that this is a more challenging dataset for resource-efficient deep learning with a lot of room for improvement.
TNN has the most remarkable performance on GTSRB dataset.
With $0.98\%$ error rate, CNN-Big model exceeds the human performance which is at $1.16\%$.

Partially discretized approaches use real-valued input which contains more information. 
Therefore, it is expected that they are able to get higher classification accuracy.
When compared to partially discretized studies, TNNs only lose a small percentage of accuracy and in return they provide better energy efficiency.
Next, we describe the unique hardware design for TNNS and investigate to which extent TNNs are area and energy efficient.

\section{Purpose-built Hardware for TNN}\label{sec:hw}
We designed a hardware architecture for TNNs that is optimized for ternary neuron weights and activation values $\{-1, 0, +1\}$.
In this section we first describe the purpose-built hardware we designed and evaluate its performance in terms of latency, throughput and energy and area efficiency.

\subsection{Hardware Architecture}

\begin{figure}[!b]
	\centering
	\includegraphics[width=\linewidth]{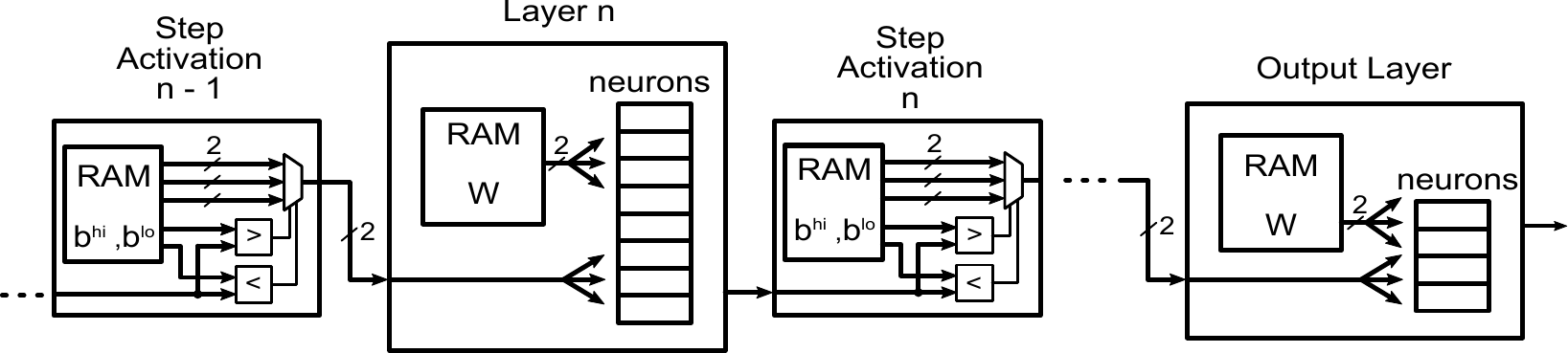}
	\caption{Hardware implementation scheme of ternary neural network}
	\label{fig:fpga-nn}
\end{figure}

Figure~\ref{fig:fpga-nn} outlines the hardware architecture of a fully-connected layer in a multi-layer NN.
The design forms a pipeline that corresponds to the sequence of NN processing steps.
For efficiency reasons, the number of layers and the maximum layer dimensions (input size and number of neurons) are decided at synthesis time.
For a given NN architecture, the design is still user-programmable: each NN layer contains a memory that can be programmed at run-time with neuron weights or output ternarization thresholds $b^{\mathit{lo}}$ and $b^{\mathit{hi}}$.
As seen in the previous experiments of Section~\ref{sec:exp}, a given NN architecture can be reused for different datasets with success.

Ternary values are represented with 2 bits using usual two's complement encoding.
That way, the compute part of each neuron is reduced to just one integer adder/subtractor and one register, both of width $\lceil\log_2 k\rceil + 1$ bits, where $k$ is the input size for the neuron.
So each neuron is only a few tens of ASIC gates, which is very small.
Inside each layer, all neurons work in parallel so that one new input item is processed per clock cycle.
Layers are pipelined in order to simultaneously work on different sets of inputs, i.e. layer $2$ processes image $n$ while layer $1$ processes image $n+1$.
The ternarization block processes the neuron outputs sequentially, so it consists of the memory of threshold values, two signed comparators and a multiplexer.

We did a generic register transfer level (RTL) implementation that can be synthesized on both Field-Programmable Gate Array (FPGA) and Application-specific Integrated Circuit (ASIC) technologies.
FPGAs are reprogrammable off-the-shelf circuits and are ideal for general-purpose hardware acceleration.
Typically, high-performance cloud solutions use high-end FPGA tightly coupled with general-purpose multicore processors~\cite{choi2016}, while ASIC is used for more throughput or in battery-powered embedded devices.

\subsection{Hardware Performance}
\label{sec:results_hw}

For the preliminary measurements, we used the dataset MNIST and the FPGA board Sakura-X~\cite{Sakura} because it features precise power measurements capabilities.
It can accommodate a 3-layer fully connected NN with 1024 neurons per layer (for a total of 3082 neurons), using 81\% of the \mbox{Kintex-7~160T} FPGA.

\begin{table}[b!]
	\centering
	\caption{FPGA Hardware performance of MLPs on Sakura-X}\label{tab:ee_perf}
	\scriptsize
	\begin{tabular}{@{}cccccc@{}}
	\toprule
	\textbf{Neurons} & \textbf{Throughput} & \textbf{Layers} & \textbf{Energy \si{\micro\joule}} &  \textbf{Latency \si{\micro\second}}  & \textbf{Accuracy} \\
					& \textbf{  (fps) }& &  \textbf{(per image)} 	& \textbf{(per image)}  & \textbf{(\%)}  \\
	\midrule
	\multirow{3}{*}{250}&\multirow{3}{*}{255 102}&1&1.24 & 5.37 & 97.76\\
	&&2&2.44 & 6.73 & 98.13\\
	&&3&3.63 & 8.09 & 98.14\\
	\midrule
	\multirow{3}{*}{500}&\multirow{3}{*}{255 102}&1&2.44 & 6.63 &97.75\\
	&&2&4.83& 9.24 &98.14\\
	&&3&7.22& 11.9  &98.29\\
	\midrule
	\multirow{3}{*}{750}&\multirow{3}{*}{255 102}&1&3.63 & 7.88 &97.73 \\
	&&2&7.22& 11.8 & 98.10\\
	&&3&10.8& 15.6  & 98.33\\
	\midrule
	\multirow{3}{*}{1000}&\multirow{3}{*}{198 019}&1&6.22 & 10.2 &97.63\\
	&&2&12.4& 15.3&98.09\\
	&&3&18.5& 20.5&97.89\\
	\bottomrule
	\end{tabular}
\end{table}

The performance of our FPGA design in terms of latency, throughput and energy efficiency is given in Table~\ref{tab:ee_perf}.
With a 200~MHz clock frequency, the throughput (here limited by the number of neurons) is \SI{195}{K} \si[per-mode=symbol]{\images\per\second} with a power consumption of \SI{3.8}{\watt} and a classification latency of \SI{20.5}{\micro\second}.

We know that TrueNorth~\cite{esser2015backpropagation} can operate at the two extremes of power consumption and accuracy.
It consumes \SI{0.268}{\micro\joule} with a network of low accuracy ($92.7\%$), and consumes as high as \SI{108}{\micro\joule} with a committee of 64 networks that achieves $99.4\%$.
Our hardware cannot operate at these two extremes, yet in the middle operating zone, we outperform TrueNorth both in terms of energy-efficiency - accuracy trade-off and speed.
TrueNorth consumes \SI{4}{\micro\joule} per image with $95\%$ accuracy with a throughput of \SI[per-mode=symbol]{1000}{\images\per\second}, and with \SI{1}{\milli\second} latency.
Our TNN hardware, consuming \SI{3.63}{\micro\joule} per image achieves $98.14\%$ accuracy at a rate of \SI[per-mode=symbol]{255102}{\images\per\second}, and a latency of \SI{8.09}{\micro\second}.

For the rest of the FPGA experiments, the larger board VC709 equipped with the \mbox{Virtex-7~690T} FPGA is used because it can support much larger designs.
We also synthesized the design as ASIC using STMicroelectronics \SI{28}{\nano\meter} FDSOI manufacturing technology.
The results are given in Table~\ref{tab:ee_perf_comp}. We compare our FPGA and ASIC solutions with the state of the art: TrueNorth~\cite{Esser2016} and EIE~\cite{EIE2016}.

\begin{table}[!t]
	\centering
	\caption{Comparison of Several Hardware Solutions for MLP}\label{tab:ee_perf_comp}
	\scriptsize
	\setlength\tabcolsep{2pt} 
	\begin{tabular}{@{}l@{}ccccc@{}}
	\toprule
	Platform & TrueNorth\cite{Esser2016} & EIE 64PE\cite{EIE2016} & EIE 256PE\cite{EIE2016} & \textbf{TNN} & \textbf{TNN}\\
	\midrule
	Year                                                           &          2014 &    2016 &             2016 &      2016 &             2016 \\
	Type                                                           &          ASIC &    ASIC &             ASIC &      FPGA &             ASIC \\
	Technology                                                     &         28 nm &   45 nm &            28 nm &  Virtex-7 &         ST 28 nm \\
	Clock (MHz)                                                    &        Async. &     800 &             1200 &       250 &              500 \\
	Quantization                                                   &         1-bit &   4-bit &            4-bit &   Ternary &          Ternary \\
	Throughput (\si{\fps})                                         &         1 989 &  81 967 & \textbf{426 230} &    61 035 &          122 070 \\
	Power (\si{\watt})                                             &          0.18 &    0.59 &             2.36 &      6.25 &            0.588 \\
	Energy Eff. (\si[per-mode=symbol]{\fps\per\watt})              &        10 839 & 138 927 &          180 606 &     9 771 & \textbf{207 567} \\
	Area (\si{\milli\squared\metre})                               &           430 &    40.8 &             63.8 &          	&    \textbf{6.36} \\
	Area Eff. (\si[per-mode=symbol]{\fps\per\milli\squared\metre}) &             5 &   2 009 &            6 681 &          	&  \textbf{19 194} \\
	\bottomrule
	\end{tabular}
\end{table}

\begin{table*}[t!]
	\centering
	\caption{Hardware performance of CNNs}\label{tab:ee_perf_cnn}
	\scriptsize
	\begin{tabular}{@{}l@{}cc@{}ccccc@{}ccccc@{}}
	\toprule
	 & \multicolumn{2}{c}{\textbf{TNN} FPGA 200~MHz} & & \multicolumn{4}{c}{\textbf{TNN} ASIC ST \SI{28}{\nano\meter} 500~MHz} & & \multicolumn{4}{c}{TrueNorth~\cite{Esser2016}}\\
	\cmidrule{2-3}\cmidrule{5-8}\cmidrule{10-13}                         & CNN-Big & CNN-Small & &         CIFAR10 &        CIFAR100 &           GTRSB &            SVNH & &  CIFAR10 &       CIFAR100 &  GTRSB &   SVNH \\
	\midrule
	Throughput (\si{\fps})                                               &   1 695 &     3 390 & &  \textbf{3 390} &  \textbf{3 390} &  \textbf{3 390} &  \textbf{6 781} & &    1 249 &          1 526 &  1 615 &  2 526 \\
	Power (\si{\watt})                                                   &    9.58 &       4.8 & &           0.377 &           0.224 &           0.310 &           0.224 & &    0.204 &          0.208 &  0.200 &  0.256 \\
	Energy per image (\si{\micro\joule})                                 &   5 650 &     1 410 & &    \textbf{111} &   \textbf{66.0} &   \textbf{91.3} &   \textbf{33.0} & &      163 &            131 &    124 &    101 \\
	Energy Efficiency (\si[per-mode=symbol]{\fps\per\watt})              &    178  &       709 & &  \textbf{8 985} & \textbf{15 148} & \textbf{10 947} & \textbf{30 344} & &    6 108 &          7 344 &  8 052 &  9 850 \\
	Area (\si{\milli\squared\metre})                                     &         &           & &   \textbf{6.06} &   \textbf{6.06} &   \textbf{6.06} &  \textbf{ 1.79} & &      424 &            424 &    424 &    424 \\
	Area Efficiency (\si[per-mode=symbol]{\fps\per\milli\squared\metre}) &         &           & &    \textbf{559} &    \textbf{559} &    \textbf{559} &  \textbf{3 787} & &     2.95 &           3.60 &   3.81 &   5.96 \\
	Accuracy (\%)                                                        &         &           & &  \textbf{87.89} &           51.60 & \textbf{ 99.02} &  \textbf{97.27} & &    83.41 & \textbf{55.64} &  96.50 &  96.66 \\
	\bottomrule
	\end{tabular}
\end{table*}

The ASIC version compares very well with TrueNorth on throughput, area efficiency (\si[per-mode=symbol]{\fps\per\milli\squared\metre}) and energy efficiency (\si[per-mode=symbol]{\fps\per\watt}).
Even though EIE uses 16-bit precision, it achieves high throughput because it takes advantage of weight sparsity and skips many useless computations.
However, we achieve better energy and area efficiencies since all our hardware elements (memories, functional units etc.) are significantly reduced thanks to ternarization.
Our energy results would be even better if taking into account weight sparsity and zero-activations (e.g. when input values are zero) like done in EIE works.

Finally, we implemented the CNN-Big and CNN-Small described in Section~\ref{sec:exp}, on both FPGA and ASIC.
Results are given in Table~\ref{tab:ee_perf_cnn}.
We give worst-case FPGA results because this is important for users of general-purpose hardware accelerators.
For ASIC technology, we took into account per-dataset zero-activations to reduce power consumption, similar to what was done in EIE works.
We compare with TrueNorth because only paper~\cite{Esser2016} gives figures of merit related to CNNs on ASIC.
The TrueNorth area is calculated according to the number of cores used.
Using different CNN models than TrueNorth's, we achieve better accuracy on three datasets out of four, while having higher throughput, better energy efficiency and much better area efficiency.

\section{Discussions and Future Work}\label{sec:concl}
In this study, we propose TNNs for resource-efficient applications of deep learning.
Energy efficiency and area efficiency are brought by not using any multiplication nor any floating-point operation.
We develop a student-teacher approach to train the TNNs and devise a purpose-built hardware for making them available for embedded applications with resource constraints.
Through experimental evaluation, we demonstrate the performance of TNNs both in terms of accuracy and resource-efficiency, with CNNs as well as MLPs.
The only other related work that has these two features is TrueNorth \cite{Esser2016}, since Bitwise NNs do not support CNNs \cite{Kim2015}.
In terms of accuracy, TNNs perform better than TrueNorth with relatively smaller networks in all of the benchmark datasets except one.
Unlike TrueNorth and Bitwise NNs, TNNs use ternary neuron activations using a step function with two thresholds.
This allows each neuron to choose a sparsity parameter for itself and gives an opportunity to remove the weights that have very little contribution.
In that respect, TNNs inherently prune the unnecessary connections.

We also develop a purpose-built hardware for TNNs that offers significant throughput and area efficiency and highly competitive energy efficiency.
As compared to TrueNorth, our TNN ASIC hardware offers improvements of 147$\times$ to 635$\times$ on area efficiency, 1.4$\times$ to 3.1$\times$ on energy efficiency and 2.1$\times$ to 2.7$\times$ on throughput.
It also often has higher accuracy with our new training approach.



\section*{Acknowledgment}
This project is being funded in part by Grenoble Alpes Métropole through the Nano2017 Esprit project.
The authors would like to thank Olivier Menut from ST Microelectronics for his valuable inputs and continuous support.

\bibliographystyle{unsrt}
\bibliography{deep}

\end{document}